\documentclass[a4paper,fleqn]{cas-sc}
\usepackage[authoryear]{natbib}
\let\cite\citep
\usepackage{amsmath,amssymb,amsfonts}
\usepackage{url}
\usepackage{graphicx}
\usepackage{xcolor}
\usepackage{multirow}
\usepackage{booktabs}
\usepackage[capitalize,noabbrev]{cleveref}

\definecolor{mygreen}{RGB}{40, 140, 70}
\newcommand{\imp}[1]{%
    \textcolor{mygreen}{%
        \sffamily\footnotesize%
        $\uparrow$#1%
    }%
}

\definecolor{myblue}{RGB}{0, 110, 180}
\newcommand{\impr}[1]{\textcolor{myblue}{\sffamily\footnotesize$\downarrow$#1}}

\begin{document}

\shorttitle{Active Spatial Guidance}
\shortauthors{Liu et~al.}

\title[mode=title]{Active Spatial Guidance: Eliminating Injected Positional Mechanisms in Vision Transformers}

\author[1]{Cong Liu}[
       type=author,
       style=normal,
       orcid=0000-0001-8624-4126]
\cormark[1]
\ead{liucong@ycit.edu.cn}
\affiliation[1]{organization={School of Engineering, Yancheng Institute of Technology},
                 addressline={No. 1 Xiwang Middle Avenue},
                 city={Yancheng},
                 postcode={224051},
                 state={Jiangsu},
                 country={PR China}}

\author[2]{Xiaofang Li}[
       type=author,
       style=normal,
       orcid=0009-0005-4857-6759]
\ead{lixf@czu.cn}
\affiliation[2]{organization={School of Computer Information Engineering, Changzhou Institute of Technology},
                 addressline={No. 666 Liaohe Road, Xinbei District},
                 city={Changzhou},
                 postcode={213032},
                 state={Jiangsu},
                 country={PR China}}

\author[3]{Simon X. Yang}[
       type=author,
       style=normal,
       orcid=0000-0002-6888-7993]
\ead{syang@uoguelph.ca}
\affiliation[3]{organization={Advanced Robotics and Intelligent Systems Laboratory, School of Engineering, University of Guelph},
                 addressline={50 Stone Road East},
                 city={Guelph},
                 postcode={N1G 2W1},
                 state={Ontario},
                 country={Canada}}

\cortext[1]{Corresponding author.}

\begin{abstract}
Vision Transformers (ViTs) commonly rely on injected positional mechanisms to address self-attention's permutation invariance. Motivated by the spatial regularities of natural images, we ask whether spatial organization can be induced from data rather than explicitly injected. Under controlled, matched from-scratch training, we propose \emph{Active Spatial Guidance} (Guidance), a training-only objective that disables positional injection and applies an auxiliary 2D coordinate-regression loss to the final-layer patch tokens. The guidance head is used only during training and removed for inference; the deployed model consists of a positional-injection-free ViT encoder and the task-specific prediction module.
Using DINOv3 ViT backbones, Guidance consistently improves performance on ImageNet-100 classification, ADE20K semantic segmentation, and Hypersim monocular depth estimation, outperforming strong injected baselines such as learned absolute positional embeddings and rotary positional embeddings under identical training protocols. On ImageNet-100, broader comparisons against representative injected positional designs further support Guidance's effectiveness. Guidance also improves robustness under resolution transfer, and multi-resolution training further strengthens accuracy across input sizes. Overall, our results suggest that spatial inductive bias in ViTs need not be architecturally injected, but can be shaped through training-time supervision. The code used for training and evaluation is publicly available in \url{https://github.com/cloudlc/asg}.
\end{abstract}


\maketitle
\section{Introduction}
\label{sec:intro}

Transformer models are built on self-attention, a permutation-invariant operator that does not, by itself, encode token order. In natural language processing, positional information is therefore essential, and positional encodings (PEs) have been a standard component since the original Transformer~\cite{vaswani_attention_2017}. Vision Transformers (ViTs)~\cite{dosovitskiy_image_2020,touvron_training_2021} largely inherited this design when adapting Transformers to images: patch tokens are paired with explicit positional signals, most commonly learned absolute embeddings and, more recently, relative biases or rotary mechanisms.

Injected positional mechanisms are effective, but they also introduce constraints in vision settings. Learned absolute positional embeddings (AbsPE) are tied to a fixed patch grid and typically require interpolation when input resolution changes, which can complicate resolution transfer and dense prediction and may reduce robustness~\cite{xie_segformer_2021,fan_vitar_2024}. These concerns have motivated alternatives that inject position through attention biases or position-dependent transformations, including relative position biases (e.g., Log-CPB), rotary embeddings (RoPE), and attention with linear biases (ALiBi)~\cite{liu_swin_2022,su_roformer_2024,press_train_2022}. Although these approaches differ in form, they share a common architectural premise: positional information is explicitly injected into the model and remains part of the forward computation at inference.

\begin{figure}
\centering
\includegraphics[width=0.96\linewidth]{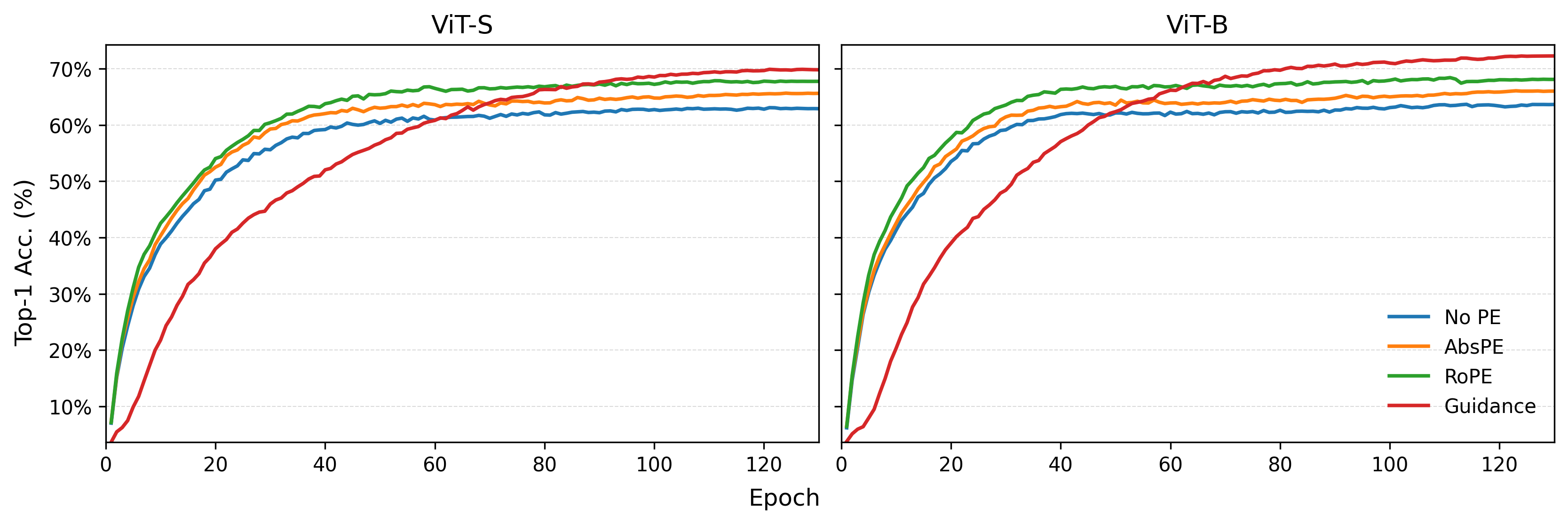}
\caption{Training dynamics on ImageNet-100 with DINOv3 ViT-S and ViT-B trained from scratch for 130 epochs at $224\times224$. Curves show validation Top-1 accuracy for a PE-free baseline (\emph{No-PE}), learned absolute positional embeddings (\emph{AbsPE}), rotary embeddings (\emph{RoPE}), and \emph{Active Spatial Guidance} (\emph{Guidance}). All settings share the same optimization and augmentation protocol; only the positional strategy differs.}
\label{fig:model_comparison_grid}
\end{figure}

At the same time, natural images exhibit strong spatial regularities: adjacent patches are highly correlated, and local continuity constrains plausible layouts. This motivates a complementary perspective: spatial representation learning may be achievable from image content and dataset structure, without an explicit positional injection mechanism. Analyses of ViT representations suggest that token features can preserve spatial information~\cite{raghu_vision_2021}. Related evidence shows that ViTs can retain non-trivial performance even when spatial structure is disrupted (e.g., via patch shuffling), indicating that explicit positional injection is not always strictly necessary for reasonable predictions~\cite{naseer_intriguing_2021}. However, standard supervised objectives do not directly require the encoder to preserve or reason about 2D patch coordinates. In our from-scratch setting (Fig.~\ref{fig:model_comparison_grid}), ViTs trained without injected positional mechanisms can converge and achieve non-trivial accuracy, yet they consistently underperform PE-equipped counterparts. This gap suggests that useful spatial structure is available in the data but is not reliably captured by a PE-free encoder under conventional training alone, motivating an explicit training-time spatial signal.

We therefore propose \emph{Active Spatial Guidance}, a simple training objective that removes injected positional mechanisms from a ViT and instead induces spatial organization through an auxiliary, training-only signal. Concretely, the guided model omits positional injection and applies an auxiliary 2D coordinate-prediction loss to final-layer patch tokens. A lightweight predictor is used only to instantiate this loss during training and is discarded afterward, leaving a PE-free ViT at inference with the same token flow and compute as the corresponding PE-free baseline.

Fig.~\ref{fig:model_comparison_grid} plots validation Top-1 accuracy over training epochs on ImageNet-100 with DINOv3 backbones~\cite{simeoni_dinov3_2025}. The PE-free baseline (\emph{No-PE}) converges but trails injected positional baselines. Among injected mechanisms, RoPE is the strongest in this from-scratch regime, improving over both \emph{No-PE} and \emph{AbsPE}. \emph{Active Spatial Guidance} starts lower early in training but steadily improves and ultimately surpasses the injected baselines for both ViT-S and ViT-B by the end of the 130-epoch budget, without using any injected positional mechanism. These dynamics suggest that, under matched training conditions, an auxiliary spatial objective can induce representations that recover and exceed the functional benefits typically provided by positional injection.

We evaluate Active Spatial Guidance under controlled from-scratch protocols on three representative tasks: image classification (ImageNet-100), semantic segmentation (ADE20K), and monocular depth estimation (Hypersim). Across tasks and model scales, we compare against PE-free baselines and strong injected positional mechanisms implemented in modern ViT backbones, including RoPE, AbsPE, ALiBi, and Log-CPB. We further study resolution transfer and multi-resolution training, and we report matched wall-clock train/validation time per epoch to assess efficiency overhead.

Our main contributions are:
\begin{itemize}
    \item We introduce \emph{Active Spatial Guidance}, a plug-and-play training objective that enables ViTs to operate without injected positional mechanisms by supervising a training-only head to regress 2D patch coordinates from final-layer patch features, yielding a PE-free inference model with no added inference-time module or compute.
    \item Through controlled from-scratch experiments on classification, semantic segmentation, and depth estimation with DINOv3 backbones, we show that the resulting PE-free models consistently outperform PE-free baselines and can surpass strong injected positional mechanisms such as RoPE, Log-CPB, and ALiBi under matched conditions.
    \item We provide robustness and efficiency analyses, including resolution transfer, multi-resolution training, matched wall-clock timing, positional-mechanism comparisons, and learning dynamics, to characterize when and why Guidance is effective.
\end{itemize}

\section{Related Work}
\label{sec:related}

Our work connects three lines of research: (i) explicit positional information mechanisms for Vision Transformers (ViTs), (ii) architectural choices that provide implicit spatial inductive bias and thereby reduce dependence on injected positional signals, and (iii) auxiliary objectives that encourage spatial reasoning. Surveys of ViTs provide broad taxonomies of architectures and positional designs~\cite{han_survey_2022}. In contrast to work that focuses on \emph{how} to inject positional information, we ask whether explicit positional injection is necessary in the deployed model, and whether spatial organization can instead be induced by a lightweight training objective.

\subsection{Positional Information Mechanisms in Vision Transformers}
Because self-attention is permutation invariant, Transformers typically require explicit positional signals to represent token order~\cite{vaswani_attention_2017}. ViTs~\cite{dosovitskiy_image_2020,touvron_training_2021} adopted the standard approach of adding learned absolute positional embeddings (AbsPE) to patch tokens, which remains a strong and widely used baseline. However, in vision, AbsPE is tied to a fixed patch grid and often requires interpolation when the input resolution changes, which can complicate resolution transfer and dense prediction and may reduce robustness~\cite{xie_segformer_2021,fan_vitar_2024}.

These limitations have motivated alternatives that inject positional information through attention biases or transformations applied to attention features (e.g., query/key rotations). Relative position bias parameterizations are widely used in vision: Swin Transformer V2 introduces a continuous position bias (CPB) computed by an MLP over relative coordinates and proposes a log-spaced variant (Log-CPB) to support stable scaling to higher resolutions and larger models~\cite{liu_swin_2022}. Rotary position embeddings (RoPE) encode positions via deterministic rotations applied to query and key vectors, enabling variable sequence lengths without explicit interpolation~\cite{su_roformer_2024,heo_rotary_2024}. Attention with linear biases (ALiBi) injects distance-dependent slopes directly into the attention logits~\cite{press_train_2022}. Beyond these general mechanisms, other work explores conditional or content-adaptive encodings~\cite{chu_conditional_2021,chen_2d_2025}, bio-inspired positional networks that learn spatial features through a convolutional pathway~\cite{tang_bio-inspired_2023}, and geometric/equivariant formulations that impose transformation structure~\cite{xu_e_2023,wang_gfpe-vit_2025}. Empirical studies have also examined what learned positional embeddings encode and how alternative positional designs affect ViT behavior~\cite{jiang_encoding_2022,wu_rethinking_2021}.

Despite their diversity, these approaches share a common architectural premise: positional information is supplied by an explicit mechanism that remains part of the model's forward computation during inference. Our work takes a complementary approach. Rather than refining positional injection, we investigate whether a standard ViT can learn spatially organized representations \emph{without} any injected positional mechanism, using a training-only coordinate-prediction objective whose auxiliary predictor is discarded after training. We distinguish this setting from architectural hybrids, discussed next, that introduce spatial bias by changing the backbone itself.

\subsection{Implicit Spatial Bias via Architectural Design}
A complementary direction reduces reliance on explicit positional mechanisms by building spatial inductive bias into the backbone itself. SegFormer~\cite{xie_segformer_2021}, for example, omits explicit positional embeddings and instead leverages overlapping patch embeddings and convolutional components (e.g., depthwise convolutions inside Mix-FFN) to promote locality and improve robustness to resolution changes. Related strategies include introducing early convolutional stages to encourage locality~\cite{xiao_early_2021}, hybrid convolution--transformer backbones that embed locality through convolutional stems or frequency-domain downsampling~\cite{wang_convolution-embedded_2022,su_dctvit_2024}, and equivariant designs that encode symmetry constraints~\cite{xu_e_2023}. Other architectural variants modify spatial interactions directly within the backbone~\cite{ma_save_2024}.

While effective, architecture-driven approaches typically introduce specialized components that change the backbone and may require additional design and tuning. In contrast, Active Spatial Guidance uses the same ViT backbone with positional injection omitted: a lightweight coordinate predictor is attached only during training and removed before inference, leaving an inference model with the same architecture and compute as the corresponding PE-free backbone.

\subsection{Auxiliary Objectives for Learning Spatial Relationships}
A third line of work encourages spatial reasoning through auxiliary objectives. Classical pretext tasks train models to predict spatial relationships between patches, such as predicting relative patch positions~\cite{doersch_unsupervised_2015} or solving jigsaw puzzles to recover patch arrangement~\cite{noroozi_mehdi_unsupervised_2016}. Transformer variants predict shuffled or masked patch positions, often using relative ordering surrogates or index-based targets~\cite{chen_jigsaw-vit_2023,zhang_positional_2023}. More recently, ViT-based jigsaw formulations have used direct fragment-coordinate regression, showing that absolute 2D coordinate recovery can be a useful supervision signal for learning spatial organization~\cite{kim_solving_2025}. These studies motivate our use of coordinate prediction, but they typically aim to solve a spatial pretext task, reconstruct shuffled layouts, or improve representation learning within a conventional ViT pipeline. They do not directly ask whether such supervision can replace injected positional mechanisms in the deployed model.

Masked Jigsaw Puzzle (MJP) position embeddings~\cite{ren_masked_2023} are closely related in spirit: MJP trains a localization regressor under patch shuffling and masking to improve robustness, while still relying on positional embeddings in the backbone. Coordinate-aware transformer designs have also used position information to guide task-specific spatial modeling, for example in anatomical landmark detection~\cite{zhu_central_2025}. In contrast, Active Spatial Guidance disables injected positional mechanisms and uses coordinate prediction purely as training-time guidance for the encoder. The supervision asks whether patch features contain enough information to recover their raster-grid locations, but the predicted coordinates are not used by the task head and the auxiliary predictor is discarded after training.

This distinction is central to our rationale. Coordinate regression is not introduced as another inference-time positional representation, nor as a stand-alone jigsaw solver. Instead, it acts as a constraint on the learned representation: if a PE-free encoder can support accurate coordinate recovery from final-layer patch tokens while also optimizing the task loss, then spatial organization has been induced in the features rather than injected into the token stream. Compared to prior jigsaw-style objectives, Active Spatial Guidance is lightweight and task-aligned: it attaches an auxiliary predictor to the final transformer layer and optimizes it jointly with the primary task objective, without introducing a separate pretraining stage. Moreover, the objective directly reflects the 2D image grid by regressing row and column coordinates, providing a direct and interpretable spatial training signal.

Finally, our work relates to observations that transformers can partially infer positional structure from other cues. In language modeling, causal masking can induce implicit positional information~\cite{irie_language_2019,haviv_transformer_2022}. In vision, ViTs without injected positional mechanisms can converge to non-trivial performance, and analyses further suggest that spatial structure and locality can emerge in practice~\cite{raghu_vision_2021,naseer_intriguing_2021}.

Overall, Active Spatial Guidance is most closely related to auxiliary spatial objectives, but differs by omitting positional injection and using coordinate recoverability purely as a training-time signal.

\section{Method}
\label{sec:method}

\begin{figure}
\centering
\includegraphics[width=1.0\textwidth]{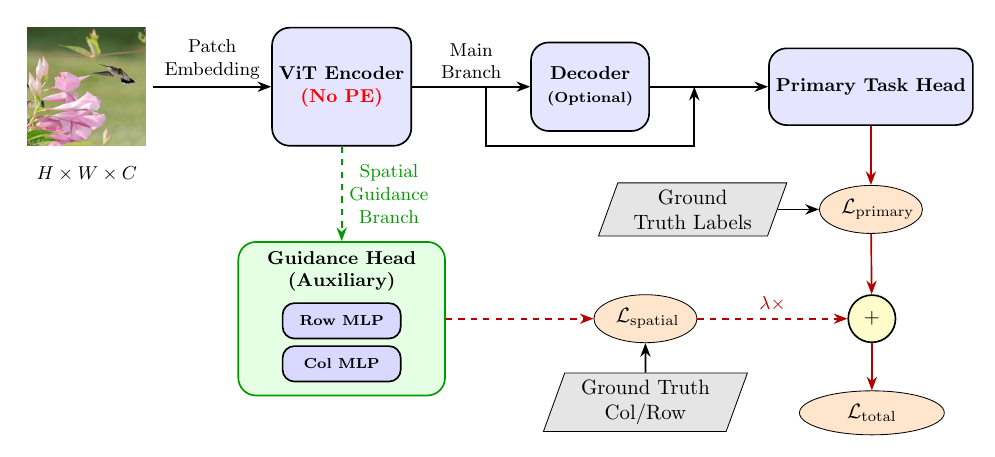}
\caption{Overview of \emph{Active Spatial Guidance}. The ViT encoder is trained with positional injection omitted (No-PE). During training, an auxiliary \emph{guidance head} attached to the final-layer patch tokens regresses normalized 2D patch-grid coordinates and provides additional spatial supervision, which is combined with the primary task loss via a scalar weight $\lambda$. The auxiliary branch is discarded after training; inference uses only the PE-free encoder and the primary task head (and an optional decoder for dense prediction).}
\label{fig:architecture_diagram}
\end{figure}

We aim to train ViTs without injected positional mechanisms while learning spatially structured representations that support strong performance across vision tasks. We propose \emph{Active Spatial Guidance} (\emph{Guidance}): (i) omit positional injection in the ViT backbone to obtain a PE-free encoder, and (ii) add a lightweight, training-only auxiliary objective that encourages final-layer patch representations to preserve the 2D organization of the patch grid. As shown in Fig.~\ref{fig:architecture_diagram}, we attach an auxiliary \emph{guidance head} to the \emph{final-layer patch tokens} and optimize a 2D coordinate-regression loss jointly with the primary task objective. The guidance head is task-independent and removed after training; therefore, inference uses only the PE-free encoder together with the primary task head (and an optional decoder for dense prediction).

\textbf{Terminology.}
We use the term positional encoding (PE) to denote any explicit mechanism that injects spatial/positional information into a ViT, including additive absolute embeddings, attention bias terms (e.g., relative/continuous biases), and rotary position-dependent transformations. \emph{Positional injection} refers to enabling such mechanisms in the backbone. \emph{No-PE} denotes an encoder in which injected positional mechanisms are absent. We use \emph{PE-free} and \emph{No-PE} interchangeably.

\subsection{Removing Positional Injection}
Given an input image of size $H\times W$, we partition it into non-overlapping patches of size $P\times P$, yielding a patch grid of size $H_p\times W_p$ and $N = H_pW_p$ patch tokens. ViT backbones commonly supply spatial information by (i) adding learned absolute positional embeddings to tokens and/or (ii) injecting position-dependent terms into attention (e.g., rotary transformations or additive/continuous biases). In the PE-free setting, we omit these positional mechanisms so that patch tokens are processed without explicit positional signals. All other components (patch embedding, special tokens such as [CLS]/register tokens, and transformer blocks) are unchanged. Concretely, we do not add absolute positional embeddings or apply RoPE rotations, and we do not instantiate any attention-bias positional terms; mechanisms such as ALiBi or (Log-)CPB are used only in their corresponding injected-position baselines.

\textbf{Absolute positional embeddings (AbsPE).}
For backbones that add learned absolute positional embeddings to patch tokens, we remove this addition. Practically, this can be implemented by bypassing the positional-embedding addition, or by setting the positional embedding parameters to zero and freezing them. Either approach preserves the remainder of the backbone while ensuring no absolute positional information is injected.

\textbf{Rotary position embeddings (RoPE).}
For backbones that apply RoPE, we disable the position-dependent transformations applied to attention queries and keys (e.g., via the corresponding RoPE flags). When disabled, attention is computed without rotary position-dependent rotations.

\subsection{Active Spatial Guidance}
Disabling injected positional mechanisms reduces the incentive for the encoder to preserve 2D structure: typical task objectives do not explicitly require patch features to make their patch-grid coordinates recoverable. Active Spatial Guidance supplies a direct training signal by requiring final-layer patch representations to predict their own locations on the patch grid, while never providing coordinates as input to the encoder. This design separates supervision from deployment: coordinates define a target that shapes the representation during training, but they are neither added to the tokens nor consumed by the inference-time model.

\subsubsection{Guidance Objective: 2D Coordinate Regression on Final-Layer Patch Tokens}
Let the output of the final transformer layer be
\[
\tilde{F}\in\mathbb{R}^{B\times S\times D},
\]
where $B$ is batch size, $D$ is embedding dimension, and $S$ is the number of emitted tokens. In general,
\[
S = N + N_{\text{sp}},
\]
where $N = H_pW_p$ is the number of patch tokens and $N_{\text{sp}}$ denotes any special tokens (e.g., class token and/or register tokens). Guidance is applied \emph{only} to patch tokens, since special tokens do not correspond to a unique 2D patch-grid location.

Let $F\in\mathbb{R}^{B\times N\times D}$ denote the final-layer patch-token tensor obtained by selecting the $N$ patch tokens from $\tilde{F}$ in the raster order produced by the patch embedding. For a patch grid of size $H_p\times W_p$, each patch token corresponds to integer coordinates $(i, j)$ with $0\le i < H_p$ and $0\le j < W_p$. We define normalized regression targets in $[0,1]$ as
\[
t^{\text{row}}_{i,j} = \frac{i}{H_p-1},\qquad
t^{\text{col}}_{i,j} = \frac{j}{W_p-1}.
\]
In our settings, $H_p, W_p > 1$, so the normalization is well-defined. Stacking targets over the grid yields per-image vectors $t^{\text{row}}, t^{\text{col}}\in\mathbb{R}^{N}$ and per-batch targets $T^{\text{row}}, T^{\text{col}}\in\mathbb{R}^{B\times N}$. These targets are deterministic functions of the patch grid and require no additional annotation. Targets are aligned with the same raster token order used to form $F$. In multi-resolution training, the targets are computed using the current patch-grid size $(H_p,W_p)$ for each batch.

\textbf{Guidance head implementation.}
The guidance head is a lightweight token-wise regressor consisting of two parallel MLPs,
\begin{align*}
\hat{T}^{\text{row}} &= g_{\text{row}}(F)\in\mathbb{R}^{B\times N},\\
\hat{T}^{\text{col}} &= g_{\text{col}}(F)\in\mathbb{R}^{B\times N},
\end{align*}
where each $g_{\cdot}$ is a two-layer MLP with hidden width 256 and a scalar output. Each MLP uses a Linear--ReLU--Linear structure, with no normalization or dropout. The guidance head is \textbf{task-independent} and shared across tasks; only the primary task head (and optional decoder) differs between classification, segmentation, and depth estimation. We attach the guidance head only to the \emph{final} transformer layer to keep auxiliary computation minimal.

We define the spatial guidance loss as the mean of the row and column regression losses over all patch tokens:
\begin{equation}
\mathcal{L}_{\text{spatial}}
=
\frac{1}{2}\Big(
\ell(\hat{T}^{\text{row}}, T^{\text{row}})
+
\ell(\hat{T}^{\text{col}}, T^{\text{col}})
\Big),
\label{eq:spatial_loss}
\end{equation}
where $\ell(\cdot,\cdot)$ is the Smooth-$\ell_1$ (Huber) loss with unit transition parameter and mean reduction (averaged over batch and patch indices). The spatial loss in Eq.~\ref{eq:spatial_loss} encourages the encoder to encode 2D patch-grid organization in its representations without injecting positional information into the token sequence.

\subsubsection{Overall Training Objective and Inference-Time Usage}
Let $\mathcal{L}_{\text{primary}}$ denote the task loss (e.g., cross-entropy for classification, pixel-wise cross-entropy for segmentation, or a depth regression loss). We optimize the composite objective
\begin{equation}
\mathcal{L}_{\text{total}}
=
\mathcal{L}_{\text{primary}}
+
\lambda\,\mathcal{L}_{\text{spatial}},
\label{eq:total_loss}
\end{equation}
where $\lambda$ controls the strength of guidance (Section~\ref{sec:experiments}). The total objective in Eq.~\ref{eq:total_loss} is optimized only during training because the guidance branch is removed before inference.

During training, gradients from both $\mathcal{L}_{\text{primary}}$ and $\mathcal{L}_{\text{spatial}}$ update the shared encoder parameters. The guidance head parameters are updated only through $\mathcal{L}_{\text{spatial}}$. At inference time, no coordinate targets are required. The deployed model therefore consists solely of the PE-free ViT encoder and the primary task head (and optional decoder), with the same inference-time architecture and compute as the corresponding \emph{No-PE} backbone.

\section{Experiments}
\label{sec:experiments}

We evaluate \emph{Active Spatial Guidance} (shorthand: \emph{Guidance}) on three representative tasks---image classification, semantic segmentation, and monocular depth estimation---using Vision Transformers trained from scratch. Unless stated otherwise, all compared models share the same backbone family, training schedule, data preprocessing, and optimization hyperparameters; configurations differ only in how spatial information is handled (injected positional mechanisms versus PE-free training with Guidance). Our goal is to isolate the effect of positional strategy under matched conditions, rather than to pursue state-of-the-art performance via large-scale pretraining or extensive task-specific tuning.

Our experiments address three questions:
\begin{enumerate}
    \item \textbf{Effectiveness:} Whether Guidance enables PE-free ViTs to match or exceed strong injected baselines under identical training protocols;
    \item \textbf{Generality:} Whether gains persist across model scales and extend to dense prediction;
    \item \textbf{Robustness and efficiency:} How Guidance behaves under resolution transfer and multi-resolution training, and whether it changes wall-clock training/validation time under matched conditions.
\end{enumerate}

\subsection{Experimental Setup}

\textbf{Backbones and positional configurations.}
Unless otherwise noted, we use DINOv3 Vision Transformers trained from scratch. For ImageNet-100 classification we evaluate ViT-S and ViT-B; for ADE20K semantic segmentation and Hypersim depth estimation we use a ViT-B encoder for consistent capacity in dense prediction.
On ImageNet-100 we compare: (i) \emph{No-PE} (PE-free backbone: positional injection removed), (ii) \emph{AbsPE} (learned absolute positional embeddings), (iii) \emph{RoPE} (rotary embeddings), and (iv) \emph{Guidance} (our training-time spatial supervision module applied to a PE-free backbone). For an extended comparison on ImageNet-100 (Section~\ref{sec:pe_compare}), we additionally evaluate injected alternatives \emph{Log-CPB} and \emph{ALiBi} under the same ViT-B training recipe.
For dense prediction, we report \emph{No-PE}, \emph{AbsPE}, \emph{RoPE}, and \emph{Guidance}, using RoPE as the primary injected reference because it is the strongest injected mechanism on ImageNet-100. In all cases, the encoder backbone and optimization protocol are held fixed; only the positional strategy differs. When Guidance is enabled, the encoder uses the No-PE configuration, and the training-only coordinate predictor and auxiliary loss are removed at inference, leaving a PE-free encoder at test time.

In DINOv3, we enable or disable positional components via \texttt{use\_abs\_pos\_emb} and \texttt{use\_rot\_pos\_emb}. For implementations where AbsPE is an additive embedding table, an equivalent way to disable AbsPE is to set the embedding weights to zero and freeze them throughout training.

\textbf{Active Spatial Guidance.}
Guidance disables injected positional mechanisms and adds an auxiliary 2D coordinate-prediction loss on the \emph{final-layer patch tokens}. A lightweight predictor instantiates this loss during training and is discarded afterward. Gradients from the guidance loss update the predictor and the shared encoder; the primary task head is optimized through the task loss.

\textbf{Optimization and schedule.}
All models are implemented in PyTorch~\cite{paszke_pytorch_2019} and trained with AdamW~\cite{loshchilov_decoupled_2019}, cosine learning rate decay, a linear warmup for 3000 optimizer steps, and FP32 precision; FP32 is used for all methods to keep numerical precision matched and make runtime comparisons directly comparable. For each task, all positional baselines and Guidance use identical hyperparameters (learning rate, weight decay, batch size, augmentations, and number of epochs). Unless stated otherwise, ImageNet-100 and Hypersim use \(224\times224\) inputs; ADE20K uses \(336\times336\). Resolution-transfer and multi-resolution settings for ImageNet-100 are described in Section~\ref{sec:res_transfer}.
All runs use weight decay \(0.01\) and global gradient clipping at \(1.0\). ImageNet-100 runs use batch size 32 (ViT-S and ViT-B) for 130 epochs; ADE20K segmentation uses batch size 16 for 130 epochs; Hypersim depth uses batch size 24 for 100 epochs. The base learning rate is \(7\times10^{-5}\) for all three tasks.

\textbf{Random seeds and reporting.}
Unless otherwise stated, we run each configuration with \textbf{four random seeds} and report results as \textbf{mean \(\pm\) standard deviation} across seeds. For each run, we report metrics from the \textbf{final training epoch} (no early stopping or best-checkpoint selection), and then aggregate across seeds. We report the final epoch to avoid confounding comparisons with checkpoint selection. We use the \textbf{same set of seeds} for all compared methods within a task to reduce variance due to initialization and data-order effects. For wide resolution-sweep tables, we report the \textbf{mean over four seeds} and omit the standard deviation for readability (std is reported for the main summary tables).

\textbf{Runtime measurement.}
To assess efficiency under matched conditions, we report runtime measurements for ImageNet-100 ViT-B under the same setting as the main classification runs. Epoch time includes data loading overhead; all timings are measured on a single GPU with the same hardware and dataloader settings as the corresponding accuracy runs. We record (i) \emph{training} time per epoch including forward, backward, and optimizer update, and (ii) \emph{validation} time per epoch using forward-only evaluation. For each configuration, we compute the median epoch time over the full training run for each seed/run, then report the mean of these medians across four seeds. All compared methods are timed with the same input resolution, batch size, and data pipeline (including augmentations) as their corresponding accuracy experiments. Guidance discards the auxiliary predictor at inference, so test-time inference uses the PE-free encoder without the guidance module.

\textbf{Guidance weight.}
We set \(\lambda\) to 100 (ViT-S) and 300 (ViT-B) for ImageNet-100, 30 for ADE20K, and 70 for Hypersim. We choose \(\lambda\) by matching early-training loss scales so that \(\lambda\,\mathcal{L}_{\text{spatial}}\) is comparable to (and slightly larger than) the primary task loss, which keeps the auxiliary signal effective even as the coordinate regression loss rapidly decreases. For each task and backbone size, we fix \(\lambda\) once and use it for all seeds and all positional comparisons. During training, we linearly warm up the guidance weight from \(\lambda_{\min}=10\) to the target \(\lambda\) over the first 60 optimizer steps (a short warmup used only to stabilize early optimization). This warmup is used because the coordinate regression loss decays quickly early in training: a large \(\lambda\) is needed to keep the auxiliary signal comparable to the primary loss, but applying that large weight from the very first steps can overly amplify the auxiliary term and destabilize early optimization.

\subsection{Task-Specific Details}

\subsubsection{Image Classification (ImageNet-100)}
\textbf{Dataset and preprocessing.}
We evaluate on \textbf{ImageNet-100}, a 100-class subset of ImageNet-1K~\cite{deng_imagenet_2009} with a standard train/validation split. Unless stated otherwise, training inputs are processed at \(224\times224\) using random resized cropping and random horizontal flipping, followed by ImageNet mean--std normalization. For validation, we resize the shorter side to \(1.143\times\) the target size and then center-crop to \(224\times224\), followed by the same normalization.

\textbf{Models and metric.}
We train DINOv3 ViT-S and ViT-B from scratch under identical schedules across positional variants and Guidance. The classification head follows the DINOv3 implementation used for all compared variants and is unchanged across methods. We report \textbf{Top-1 accuracy} on the ImageNet-100 validation set.

\textbf{Extended injected positional mechanisms.}
\label{sec:pe_compare}
To contextualize Guidance against a broader family of injected designs, we additionally evaluate ViT-B with representative injected mechanisms under the same ImageNet-100 training recipe: learned absolute positional embeddings (AbsPE), rotary embeddings (RoPE), ALiBi, and Log-CPB.

\textbf{Resolution transfer and multi-resolution training.}
\label{sec:res_transfer}
To assess robustness to input size changes, we evaluate Top-1 accuracy over a sweep of test resolutions while varying the training regime. \textbf{SINGLE-RES} trains at \(224\times224\) only, while \textbf{MULTI-RES} samples training resolutions from \(\{192, 224, 288\}\) on a per-batch basis. All models are evaluated on the ImageNet-100 validation set at each test resolution. For compactness in the resolution-sweep table, we report means over four seeds and omit standard deviations.

\subsubsection{Semantic Segmentation (ADE20K)}
\textbf{Dataset and preprocessing.}
We evaluate on \textbf{ADE20K}~\cite{zhou_scene_2017}, which contains 150 semantic categories, using the official train/validation split. Training uses paired augmentations with scale jitter in \([1.0, 1.3]\), category-max-ratio sampling (ratio \(0.70\), 10 retries), and color jitter (brightness/contrast/saturation \(0.2\), hue \(0.05\)) applied with probability 0.1; images are normalized with ImageNet mean--std. Both training and validation use a target size of \(336\times336\). For validation, we resize by the shorter side and then crop-or-pad to the target size. Masks are shifted by \(-1\) so that labels map to \([0, 149]\) and background becomes \(-1\), which is ignored by the loss.

\textbf{Model, loss, and metrics.}
We use a ViT-B encoder and a UPerNet-style token decoder~\cite{xiao_unified_2018} that fuses multi-level features (layers \(\{2, 5, 8, 11\}\)) with an FPN and a PPM (bins \(\{1, 2, 3, 6\}\)), using 256 FPN channels, GroupNorm, and 0.1 dropout; logits are upsampled using bilinear interpolation with \texttt{align\_corners} disabled. All variants share the same encoder--decoder architecture; only the encoder positional strategy differs, and when Guidance is enabled, positional injection is omitted in the encoder. The primary loss is pixel-wise cross-entropy, excluding unlabeled pixels using \texttt{ignore\_index}=$-1$ and averaging over non-ignored pixels. We report \textbf{pixel accuracy} and \textbf{mIoU} on the validation set under single-scale, no-flip evaluation.

\subsubsection{Monocular Depth Estimation (Hypersim)}
\textbf{Dataset and preprocessing.}
We evaluate monocular depth estimation on \textbf{Hypersim}~\cite{roberts_hypersim_2021} using the official train/validation split. Training uses paired augmentations with scale jitter in \([1.0, 1.2]\) and color jitter (probability 0.5), followed by normalization and resizing to \(224\times224\). Validation resizes the shorter side to 256 and then center-crops to \(224\times224\), followed by normalization.

\textbf{Model, losses, and metrics.}
Our depth model consists of a ViT-B encoder and a DepthAnything-style DPT decoder~\cite{yang2024depth}, which consumes multi-level transformer features (layers \(\{2, 5, 8, 11\}\)) and produces a single-channel depth map. During training, we optimize a hybrid loss composed of per-pixel \(L_1\) and gradient terms (\(1.0\times L_1\) and \(0.5\times\) gradient loss), computed over valid pixels (depth \(>\) 0). At evaluation, we use relative-depth alignment via scale-and-shift fitting and clamp predictions to the valid depth range before computing metrics. We report \textbf{AbsRel} and \(a_1\) on the validation set.

\section{Results}
\label{sec:results}

We evaluate \emph{Active Spatial Guidance} (\emph{Guidance}) under controlled, from-scratch training conditions. Unless otherwise stated, all compared models use the same DINOv3 ViT backbone family, data preprocessing, augmentation, and optimization hyperparameters; configurations differ only in how spatial information is handled (injected positional mechanisms versus PE-free training with Guidance). We first report ImageNet-100 classification results and place Guidance in context by comparing against additional injected positional mechanisms. We then test whether the gains extend beyond classification by evaluating semantic segmentation on ADE20K and monocular depth estimation on Hypersim. Next, we analyze resolution transfer under single- and multi-resolution training. Finally, we report matched runtime measurements to determine whether Guidance introduces measurable efficiency overhead.

\subsection{Main Classification Results}
\label{sec:results_main_cls}

\begin{table}[t]
\centering
\setlength{\tabcolsep}{6pt}
\caption{ImageNet-100 Top-1 accuracy (\%) for DINOv3 ViT-S and ViT-B trained from scratch for 130 epochs at $224\times224$. Results are reported as mean$\pm$std over \textbf{four seeds}. \textbf{Guidance} denotes our training-time 2D coordinate guidance (removed at inference). \emph{No-PE}: no injected positional mechanism; \emph{AbsPE}: learned absolute positional embeddings; \emph{RoPE}: rotary positional embeddings. RoPE (marked by $^\dagger$) is the reference for \imp{} (absolute gain over RoPE).}
\label{tab:main_imagenet100_dinov3}
\setlength{\tabcolsep}{3pt}
\begin{tabular}{lcccc}
\toprule
Model & No-PE & AbsPE~\cite{dosovitskiy_image_2020} & RoPE$^\dagger$~\cite{su_roformer_2024} & \textbf{Guidance} \\
\midrule
ViT-S & 62.89$\pm$0.51 & 65.61$\pm$0.79 & 67.74$\pm$0.62 & \textbf{69.80$\pm$0.84}~\imp{2.06} \\
ViT-B & 63.64$\pm$0.10 & 66.01$\pm$0.27 & 68.09$\pm$0.37 & \textbf{72.26$\pm$0.70}~\imp{4.17} \\
\bottomrule
\end{tabular}
\end{table}

Table~\ref{tab:main_imagenet100_dinov3} reports ImageNet-100 Top-1 accuracy for ViT-S and ViT-B trained from scratch under a fixed 130-epoch protocol. Disabling positional injection (\emph{No-PE}) yields 62.89\% (ViT-S) and 63.64\% (ViT-B), corresponding to drops of 4.85 and 4.45 points relative to RoPE (67.74\%/68.09\%). Among injected baselines, learned absolute positional embeddings (\emph{AbsPE}) improve performance over \emph{No-PE} but remain below \emph{RoPE} (by 2.13 points for ViT-S and 2.08 points for ViT-B). We therefore use RoPE as the primary injected baseline in the remainder of the paper.

\emph{Guidance} achieves the highest accuracy while remaining PE-free at inference time. Guidance reaches 69.80\% on ViT-S and 72.26\% on ViT-B, outperforming RoPE by \imp{2.06} and \imp{4.17} points, respectively (Table~\ref{tab:main_imagenet100_dinov3}). The learning curves in Fig.~\ref{fig:model_comparison_grid} further show that Guidance closes an early gap and surpasses injected baselines within the 130-epoch budget. Overall, these results indicate that training-time coordinate guidance can improve PE-free representations for recognition without injecting positional mechanisms at test time.

\subsection{Comparison with Injected Positional Mechanisms}
\label{sec:results_mechanism}

This subsection situates Guidance relative to a broader set of injected positional mechanisms on ImageNet-100. Using ViT-B trained from scratch for 130 epochs at $224\times224$, Table~\ref{tab:comparison_methods} reports results for AbsPE, RoPE, Log-CPB, and ALiBi, alongside the PE-free \emph{No-PE} baseline.
All injected mechanisms improve over \emph{No-PE}, indicating that explicit positional designs provide a beneficial spatial inductive bias in this from-scratch setting. All mechanisms use the same training recipe and their default implementation settings; we do not perform mechanism-specific hyperparameter tuning. Among the injected mechanisms evaluated, RoPE is strongest (68.09\%), followed closely by Log-CPB (67.82\%); AbsPE (66.01\%) and ALiBi (65.94\%) yield smaller gains. Guidance achieves the highest accuracy (72.26\%), corresponding to +8.62 points over \emph{No-PE} and +4.17 points over the best injected mechanism in this comparison (RoPE). These results show that, under this matched recipe, replacing positional injection with training-time coordinate guidance is more effective than the injected mechanisms evaluated here.

\begin{table}[t]
\centering
\caption{Comparison of strategies for supplying spatial information on ImageNet-100 (ViT-B), trained from scratch for 130 epochs at $224\times224$. Results are reported as mean$\pm$std over \textbf{four seeds}. \emph{Log-CPB} denotes log-spaced continuous position bias.}
\label{tab:comparison_methods}
\begin{tabular}{lc}
\toprule
\textbf{Method} & \textbf{Top-1 Acc. (\%)} \\
\midrule
\textbf{Guidance} & \textbf{72.26$\pm$0.70} \\
RoPE~\cite{su_roformer_2024}            & 68.09$\pm$0.37 \\
Log-CPB~\cite{liu_swin_2022}            & 67.82$\pm$0.67 \\
AbsPE~\cite{dosovitskiy_image_2020}     & 66.01$\pm$0.27 \\
ALiBi~\cite{press_train_2022}           & 65.94$\pm$0.29 \\
No-PE                                  & 63.64$\pm$0.10 \\
\bottomrule
\end{tabular}
\end{table}

\subsection{Beyond Classification}
\label{sec:results_beyond_cls}

We next evaluate whether the benefits of Guidance extend to dense prediction tasks under matched training conditions, varying the encoder positional strategy and, for Guidance, adding the training-only guidance branch.

\subsubsection{Semantic Segmentation on ADE20K}
Table~\ref{tab:segmentation_performance_revised} reports ADE20K results for a ViT-B encoder trained from scratch with a UPerNet-style token decoder at $336\times336$. Removing positional injection (\emph{No-PE}) yields the weakest performance (64.20\% pixel accuracy and 17.36\% mIoU), consistent with the challenge of learning spatial correspondence without an explicit spatial bias. Injected baselines (AbsPE and RoPE) improve both metrics. Guidance yields further gains, reaching 67.06\% pixel accuracy and 19.85\% mIoU, improving over RoPE by \imp{1.72} and \imp{1.86} points, respectively. These improvements are achieved while keeping the \emph{encoder} PE-free at inference time, with no injected positional mechanism and with the guidance predictor removed after training.

\begin{table}[htbp]
\centering
\caption{ADE20K semantic segmentation results for ViT-B trained from scratch at $336\times336$, reported as mean$\pm$std over \textbf{four seeds}. \textbf{Guidance} denotes Active Spatial Guidance. Acc.\ denotes pixel accuracy; \imp{} is absolute gain over RoPE$^\dagger$.}
\label{tab:segmentation_performance_revised}
\setlength{\tabcolsep}{4pt}
\begin{tabular}{lcccc}
\toprule
\textbf{Metric} & \textbf{No-PE} & \textbf{AbsPE} & \textbf{RoPE}$^\dagger$ & \textbf{Guidance} \\
\midrule
Acc. (\%) & 64.20$\pm$0.09 & 64.92$\pm$0.19 & 65.34$\pm$0.18 & \textbf{67.06$\pm$0.11}~\imp{1.72} \\
mIoU (\%) & 17.36$\pm$0.14 & 17.91$\pm$0.12 & 17.99$\pm$0.15 & \textbf{19.85$\pm$0.12}~\imp{1.86} \\
\bottomrule
\end{tabular}
\end{table}

\subsubsection{Monocular Depth Estimation on Hypersim}
Table~\ref{tab:depth_estimation_hypersim} reports Hypersim results using AbsRel (\(\downarrow\)) and \(a_1\) (\(\uparrow\)). We report AbsRel as \(\text{AbsRel}\times 100\) for readability and \(a_1\) in \%. The PE-free baseline (\emph{No-PE}) performs worst (AbsRel 28.15, \(a_1\) 65.23\%), while RoPE improves both metrics (AbsRel 27.28, \(a_1\) 65.55\%). Guidance achieves the best performance, reducing AbsRel to 25.35 (\impr{1.93} vs.\ RoPE) and increasing \(a_1\) to 68.77\% (\imp{3.22} vs.\ RoPE). Together with the segmentation results above, these findings indicate that the same training-only coordinate objective that improves classification also yields gains on dense semantic and geometric prediction.

\begin{table}[t]
\centering
\caption{Hypersim monocular depth estimation results with a ViT-B encoder trained from scratch, reported as mean$\pm$std over \textbf{four seeds}. AbsRel is reported as \(\text{AbsRel}\times 100\) (lower is better) and \(a_1\) is reported in \% (higher is better). RoPE$^\dagger$ is the reference baseline for \impr{} (absolute reduction in AbsRel$\times 100$) and \imp{} (absolute gain in \(a_1\)).}
\label{tab:depth_estimation_hypersim}
\setlength{\tabcolsep}{8pt}
\begin{tabular}{lcc}
\toprule
\textbf{Method} & \textbf{AbsRel$\times 100$} \(\downarrow\) & \(\mathbf{a_1}\) \textbf{(\%)} \(\uparrow\) \\
\midrule
No-PE             & 28.15$\pm$0.18 & 65.23$\pm$0.24 \\
AbsPE             & 28.07$\pm$0.08 & 65.44$\pm$0.23 \\
RoPE$^\dagger$    & 27.28$\pm$0.19 & 65.55$\pm$0.20 \\
\textbf{Guidance} & \textbf{25.35$\pm$0.36}~\impr{1.93} & \textbf{68.77$\pm$0.39}~\imp{3.22} \\
\bottomrule
\end{tabular}
\end{table}

\subsection{Resolution Robustness and Multi-Resolution Training}
\label{sec:results_res_transfer}

We evaluate resolution transfer by training models at one or more resolutions and testing at a sweep of image sizes. This probes whether the learned spatial organization generalizes across patch-grid sizes and whether different positional strategies exhibit sensitivity to resolution changes.

Table~\ref{tab:res_transfer_transposed_single_new} reports Top-1 accuracy across evaluation resolutions for two training regimes. In \textbf{SINGLE-RES}, models are trained only at $224\times224$ and evaluated from 160 to 416 pixels. In \textbf{MULTI-RES}, training samples resolutions from $\{192, 224, 288\}$ on a per-batch basis, and evaluation uses the same resolution sweep.

Under \textbf{SINGLE-RES} training, Guidance achieves the highest accuracy at most evaluation resolutions. At the nominal training resolution (224), Guidance reaches 72.26\% versus 68.09\% (RoPE) and 66.01\% (AbsPE). Guidance also maintains an advantage across lower and intermediate resolutions (e.g., 192: 68.96\% vs.\ 65.36\% for RoPE; 256: 73.72\% vs.\ 69.88\%; 320: 72.80\% vs.\ 71.27\%). At the highest tested resolution (416), RoPE is marginally higher (70.17\% vs.\ 70.04\%), while Guidance remains competitive overall.

Although Guidance often achieves the best absolute accuracy, its \emph{drop} under large resolution shifts can be larger when trained at a single resolution. For example, Guidance decreases from 72.26\% at 224 to 60.11\% at 160 ($-12.15$ points), whereas RoPE decreases from 68.09\% to 60.05\% ($-8.04$ points). Despite the larger drop, Guidance still matches RoPE at the lowest tested resolution (160). This suggests that, when trained at a single resolution, the spatial calibration induced by the coordinate objective may partially specialize to the training patch lattice.

Multi-resolution training strengthens resolution transfer, with the largest gains observed for Guidance. Compared to SINGLE-RES, \textbf{MULTI-RES} substantially improves Guidance performance at off-training resolutions, particularly at higher input sizes (e.g., 416: 77.38\% vs.\ 70.04\%, a $+7.34$ point gain). At 224, Guidance changes minimally (72.20\% vs.\ 72.26\%). Guidance also preserves a large margin over RoPE under MULTI-RES across the evaluation sweep (e.g., 224: 72.20\% vs.\ 67.54\%; 416: 77.38\% vs.\ 70.81\%). Overall, these results suggest that exposing Guidance to multiple patch-grid sizes during training improves the resolution generality of the learned spatial structure.

\begin{table*}[htbp]
\centering
\setlength{\tabcolsep}{3pt}
\caption{Top-1 accuracy (\%) under resolution transfer on ImageNet-100 with DINOv3 ViT-B trained from scratch. \textbf{SINGLE-RES} trains at $224\times224$; \textbf{MULTI-RES} samples training resolutions from $192\times192/224\times224/288\times288$ (boldfaced in the \emph{Res.} row). \emph{Res.} denotes the evaluation input \textbf{resolution}. Values are reported as \textbf{mean over four seeds} (std omitted for readability). \textbf{Guidance}: Active Spatial Guidance. \textbf{RoPE}/\textbf{AbsPE}/\textbf{No-PE}: rotary positional embeddings / learned absolute positional embeddings / no injected positional mechanism.}
\label{tab:res_transfer_transposed_single_new}
\resizebox{\textwidth}{!}{%
\begin{tabular}{llccccccccccccccc}
\toprule
\multirow{6}{*}{\rotatebox{90}{{SINGLE-RES}}}
& Res. (px) & 160 & 176 & 192 & 208 & \textbf{224} & 256 & 272 & 288 & 320 & 336 & 352 & 368 & 384 & 400 & 416 \\
\cmidrule(lr){2-17}
& No-PE    & 56.57 & 59.36 & 60.59 & 62.55 & 63.64 & 64.97 & 65.23 & 65.69 & 66.40 & 66.50 & 66.32 & 66.30 & 66.41 & 65.97 & 65.91 \\
& AbsPE    & 58.90 & 62.40 & 63.61 & 65.22 & 66.01 & 66.70 & 67.12 & 67.80 & 68.18 & 68.26 & 68.25 & 68.01 & 67.91 & 67.66 & 67.49 \\
& RoPE     & 60.05 & 63.45 & 65.36 & 67.03 & 68.09 & 69.88 & 70.71 & 70.99 & 71.27 & 71.15 & 71.00 & 70.89 & 70.92 & 70.50 & 70.17 \\
& Guidance & 60.11 & 65.44 & 68.96 & 70.87 & 72.26 & 73.72 & 73.19 & 73.33 & 72.80 & 71.92 & 71.89 & 71.80 & 71.40 & 70.82 & 70.04 \\
\specialrule{0.7pt}{2pt}{3pt}
\multirow{4}{*}{\rotatebox{90}{{MULTI-RES}}}
& Res. (px) & 160 & 176 & \textbf{192} & 208 & \textbf{224} & 256 & 272 & \textbf{288} & 320 & 336 & 352 & 368 & 384 & 400 & 416 \\
\cmidrule(lr){2-17}
& No-PE    & 57.50 & 60.10 & 61.29 & 62.44 & 63.41 & 64.94 & 65.19 & 65.74 & 65.93 & 66.61 & 66.79 & 66.70 & 67.11 & 67.26 & 67.14 \\
& AbsPE    & 59.89 & 62.05 & 63.71 & 64.62 & 65.44 & 66.89 & 67.25 & 67.20 & 68.03 & 68.00 & 68.36 & 68.13 & 68.36 & 68.34 & 68.65 \\
& RoPE     & 62.03 & 64.41 & 65.87 & 66.64 & 67.54 & 69.15 & 69.63 & 69.87 & 70.50 & 70.39 & 70.60 & 70.65 & 71.01 & 71.00 & 70.81 \\
& Guidance & 67.98 & 69.38 & 70.26 & 72.02 & 72.20 & 74.24 & 74.30 & 75.08 & 76.02 & 76.10 & 77.28 & 77.18 & 77.02 & 77.22 & 77.38 \\
\bottomrule
\end{tabular}%
}
\end{table*}

\subsection{Runtime and Efficiency}
\label{sec:results_runtime}

We next examine whether Guidance incurs measurable runtime overhead under the same training and evaluation pipeline. Table~\ref{tab:epoch_time_speed} reports the \textbf{mean of per-run median wall-clock time per epoch} for training and validation (four seeds per configuration) measured on a single GPU. Overhead $\Delta$ (\%) is reported relative to the \emph{No-PE} baseline within the same model size.

Across both ViT-B and ViT-S, the validation times of \emph{No-PE}, \emph{AbsPE}, and \textbf{Guidance} are nearly identical (within $\approx$1.5\% for ViT-B and within $\approx$0.5\% for ViT-S), and the small differences should be interpreted as normal run-to-run timing variability rather than a systematic efficiency effect. Training times for \emph{No-PE} and \emph{AbsPE} are also essentially unchanged. Guidance adds a lightweight auxiliary prediction head only during training; its measured overhead remains about 1\% or less for ViT-B (+0.6\%) and ViT-S (+1.0\%), and the predictor is removed after training.

In contrast, RoPE introduces a clear and consistent overhead in both training and validation across model sizes due to additional position-dependent computations in attention, increasing per-epoch time relative to \emph{No-PE} by +9.6\%/+14.4\% (train/val) for ViT-B and +14.8\%/+15.3\% for ViT-S (Table~\ref{tab:epoch_time_speed}). Overall, these results indicate that Guidance delivers the accuracy and robustness gains reported above without a meaningful runtime penalty in this setup. Moreover, since the Guidance predictor is discarded at inference, the inference-time encoder matches the corresponding \emph{No-PE} encoder in both architecture and compute.

\begin{table}[t]
\centering
\caption{Per-epoch wall-clock time (seconds) for training and validation on ImageNet-100 measured on a single GPU. For each seed/run, we compute the median epoch time over the full training run, then report the mean of these per-run medians across \textbf{four seeds}. Overhead $\Delta$ (\%) is computed relative to the No-PE baseline within the same model size. Training includes forward+backward+optimizer; validation is forward-only. Small overhead values near 0\% should be interpreted as run-to-run timing variability rather than a systematic efficiency difference.}
\label{tab:epoch_time_speed}
\setlength{\tabcolsep}{6pt}
\begin{tabular}{llcccc}
\toprule
Model & Method & Train & $\Delta$ (\%) & Val & $\Delta$ (\%) \\
\midrule
\multirow{6}{*}{ViT-B}
 & No-PE        & 2373.2 & --  & 28.35 & --  \\
 & AbsPE        & 2384.0 & +0.5  & 28.75 & +1.4  \\
 & RoPE         & 2601.1 & +9.6  & 32.44 & +14.4 \\
 & Log-CPB      & 2420.4 & +2.0  & 28.78 & +1.5  \\
 & ALiBi        & 2423.3 & +2.1  & 30.02 & +5.9  \\
 & \textbf{Guidance}
               & 2388.5 & +0.6  & 28.46 & +0.4  \\
\midrule
\multirow{4}{*}{ViT-S}
 & No-PE        & 710.3  & --  & 16.05 & --  \\
 & AbsPE        & 711.4  & +0.2  & 16.07 & +0.1  \\
 & RoPE         & 815.8  & +14.8 & 18.50 & +15.3 \\
 & \textbf{Guidance}
               & 717.7  & +1.0  & 15.99 & $-$0.4 \\
\bottomrule
\end{tabular}
\end{table}

\section{Discussion: Guidance-Driven Spatial Learning vs. Positional Encodings}
\label{sec:discussion}

Across our controlled comparisons, a consistent pattern emerges: under matched, from-scratch training with DINOv3 ViTs, the best measured performance is achieved not by injecting positional metadata (AbsPE, RoPE, Log-CPB, ALiBi), but by \emph{inducing} spatial organization through \emph{Active Spatial Guidance}. A PE-free encoder (\emph{No-PE}) can optimize and converge, yet it lags behind PE-equipped counterparts, indicating that standard task supervision does not reliably compel the backbone to preserve 2D structure. Injected positional mechanisms partially close this gap by supplying location information as an architectural signal. However, under the same backbones and optimization protocol, Guidance surpasses the injected alternatives evaluated here across model scales and across both recognition and dense prediction tasks.

This contrast is best understood as \emph{providing} position versus \emph{training} the representation to make position recoverable. Injected mechanisms provide positional signals through a predefined parameterization that is present throughout training and remains part of the inference-time model. In Guidance, the encoder is kept PE-free: spatial structure is encouraged only through an auxiliary objective that asks final-layer patch features to predict their normalized 2D grid coordinates. The lightweight predictor exists solely to instantiate this loss. The substantive change is the training signal: the encoder is rewarded when its representations make spatial coordinates inferable from content-driven features and interactions. Because the predictor is removed after training, the inference-time model is the corresponding PE-free ViT coupled with the task head, with no positional module or guidance branch.

The resolution-transfer results refine this picture. Under SINGLE-RES training, Guidance attains the highest accuracy at most evaluation resolutions, but it can be more sensitive to large resolution shifts than injected baselines, consistent with some specialization of the learned spatial calibration to the training patch lattice when only one grid is observed during optimization. Under MULTI-RES training, this sensitivity is substantially mitigated: exposing the model to multiple patch-grid sizes during training yields a stronger transfer profile, and Guidance shows the largest robustness gains among the compared methods. Overall, these results suggest complementarity between (i) varying the patch grid during optimization and (ii) explicitly rewarding representations that preserve 2D location information.

Efficiency considerations further motivate guidance-style training for practical vision systems. RoPE operates inside the attention computation of every transformer layer by applying position-dependent transformations to query/key features; consequently, any added cost is incurred repeatedly within the dominant attention blocks and tends to increase with model depth. In contrast, AbsPE is applied once at the input as an additive embedding, and Active Spatial Guidance introduces only a lightweight auxiliary head during training and removes it at inference; in our wall-clock train/validation time-per-epoch measurements, Guidance adds minimal overhead relative to \emph{No-PE}, while RoPE introduces a clear per-epoch runtime increase. This distinction is relevant in settings where throughput and latency constraints are important alongside accuracy.

Finally, we emphasize the scope of the study. We intentionally focus on from-scratch training with DINOv3 backbones, moderate input resolutions, and matched architectures and optimization settings to isolate the role of positional strategy. Within this controlled regime, the results support a reframing: spatial representation learning in ViTs can be shaped by a task-agnostic training objective, rather than only by metadata engineered and injected into the encoder. In this sense, our findings complement recent work studying positional parameterizations and their effects on training dynamics and robustness in vision transformers~\cite{han_survey_2022,jiang_encoding_2022}.

\section{Conclusion and Future Work}
\label{sec:conclusion}

We revisited the common assumption that Vision Transformers require injected positional mechanisms for effective learning. Under matched, from-scratch training with DINOv3 ViTs, we find that while injected positional mechanisms improve over a PE-free baseline, they are not the most effective route among the evaluated strategies in our setting. We introduced \emph{Active Spatial Guidance}, a plug-and-play training objective for PE-free ViTs that applies a training-only 2D coordinate-prediction loss to final-layer patch tokens. Across ImageNet-100 classification, ADE20K semantic segmentation, and Hypersim monocular depth estimation, Guidance consistently outperforms strong injected baselines under identical backbones and optimization protocols and improves robustness under resolution transfer, particularly when combined with multi-resolution training. Because the auxiliary predictor is discarded after training, the inference-time model remains the corresponding PE-free ViT with no inference-time overhead relative to the PE-free backbone, which can be attractive when simplicity and efficiency are important. The present study is limited to from-scratch training with DINOv3 backbones, moderate input resolutions, and the evaluated classification, segmentation, and depth-estimation benchmarks; its behavior under large-scale pretraining, broader architectures, and additional downstream tasks remains to be established.

\textbf{Future Work.}
A key next step is to test Guidance in large-scale self-supervised or masked-image pretraining and quantify downstream transfer gains under standard fine-tuning. Another direction is to extend the same training principle to other grid-structured modalities where positional injection is common (e.g., volumetric data or time--frequency representations) and to study when training-time coordinate recovery can replace or complement injected positional metadata.

\section*{Data availability statement}
The datasets used in this study are publicly available: ImageNet-100
(\url{https://www.kaggle.com/datasets/ambityga/imagenet100}), ADE20K
(\url{https://www.kaggle.com/datasets/awsaf49/ade20k-dataset}), and Hypersim
(\url{https://github.com/apple/ml-hypersim}).

\bibliographystyle{cas-model2-names}
\bibliography{refs}

@inproceedings{xie_segformer_2021,
	title = {{SegFormer}: {Simple} and {Efficient} {Design} for {Semantic} {Segmentation} with {Transformers}},
	volume = {34},
	shorttitle = {{SegFormer}},

	booktitle = {Advances in {Neural} {Information} {Processing} {Systems}},
	author = {Xie, Enze and Wang, Wenhai and Yu, Zhiding and Anandkumar, Anima and Alvarez, Jose M. and Luo, Ping},
	year = {2021},
	pages = {12077--12090},

}

@inproceedings{irie_language_2019,
	title = {Language {Modeling} with {Deep} {Transformers}},
	doi = {10.21437/Interspeech.2019-2225},
	booktitle = {Interspeech 2019},
	author = {Irie, Kazuki and Zeyer, Albert and Schlüter, Ralf and Ney, Hermann},
	month = sep,
	year = {2019},
	pages = {3905--3909},
}

@misc{fan_vitar_2024,
	title = {{ViTAR}: {Vision} {Transformer} with {Any} {Resolution}},
	shorttitle = {{ViTAR}},
	doi = {10.48550/arXiv.2403.18361},
	author = {Fan, Qihang and You, Quanzeng and Han, Xiaotian and Liu, Yongfei and Tao, Yunzhe and Huang, Huaibo and He, Ran and Yang, Hongxia},
	month = mar,
	year = {2024},
	journal = {arXiv preprint arXiv:2403.18361},
}

@article{ma_save_2024,
	title = {{SAVE}: {Encoding} spatial interactions for vision transformers},
	volume = {152},
	issn = {0262-8856},
	shorttitle = {{SAVE}},
	doi = {10.1016/j.imavis.2024.105312},

	journal = {Image and Vision Computing},
	author = {Ma, Xiao and Zhang, Zetian and Yu, Rong and Ji, Zexuan and Li, Mingchao and Zhang, Yuhan and Chen, Qiang},
	month = dec,
	year = {2024},
	pages = {Article 105312},
}

@misc{chen_2d_2025,
	title = {A {2D} {Semantic}-{Aware} {Position} {Encoding} for {Vision} {Transformers}},
	doi = {10.48550/arXiv.2505.09466},
	author = {Chen, Xi and Zhou, Shiyang and Huang, Muqi and Feng, Jiaxu and Xiong, Yun and Zhou, Kun and Yang, Biao and Zhang, Yuhui and Bao, Huishuai and Peng, Sijia and Li, Chuan and Shi, Feng},
	month = may,
	year = {2025},
	journal = {arXiv preprint arXiv:2505.09466},
}

@article{wang_convolution-embedded_2022,
	title = {Convolution-{Embedded} {Vision} {Transformer} {With} {Elastic} {Positional} {Encoding} for {Pansharpening}},
	volume = {60},
	issn = {1558-0644},
	doi = {10.1109/TGRS.2022.3227405},
	journal = {IEEE Transactions on Geoscience and Remote Sensing},
	author = {Wang, Nan and Meng, Xiangjun and Meng, Xiangchao and Shao, Feng},
	year = {2022},
	pages = {1--9},
}

@article{su_roformer_2024,
	title = {{RoFormer}: {Enhanced} transformer with {Rotary} {Position} {Embedding}},
	volume = {568},
	issn = {0925-2312},
	shorttitle = {{RoFormer}},
	doi = {10.1016/j.neucom.2023.127063},

	journal = {Neurocomputing},
	author = {Su, Jianlin and Ahmed, Murtadha and Lu, Yu and Pan, Shengfeng and Bo, Wen and Liu, Yunfeng},
	month = feb,
	year = {2024},
	pages = {Article 127063},

}

@inproceedings{vaswani_attention_2017,
	title = {Attention is {All} you {Need}},
	volume = {30},

	booktitle = {Advances in {Neural} {Information} {Processing} {Systems}},
	author = {Vaswani, Ashish and Shazeer, Noam and Parmar, Niki and Uszkoreit, Jakob and Jones, Llion and Gomez, Aidan N and Kaiser, {\L}ukasz and Polosukhin, Illia},
	year = {2017},

	pages = {5998--6008},
}

@inproceedings{noroozi_mehdi_unsupervised_2016,
	title = {Unsupervised {Learning} of {Visual} {Representations} by {Solving} {Jigsaw} {Puzzles}},
	booktitle = {European conference on computer vision},
	publisher = {Springer},
	author = {Noroozi, Mehdi and Favaro, Paolo},
	year = {2016},
	pages = {69--84},

}

@article{wang_gfpe-vit_2025,
	title = {{GFPE}-{ViT}: vision transformer with geometric-fractal-based position encoding},
	volume = {41},
	issn = {1432-2315},
	shorttitle = {{GFPE}-{ViT}},
	doi = {10.1007/s00371-024-03381-8},

	language = {en},
	number = {2},
	journal = {Vis Comput},
	author = {Wang, Lei and Tang, Xue-song and Hao, Kuangrong},
	month = jan,
	year = {2025},
	pages = {1021--1036},
}

@misc{simeoni_dinov3_2025,
	title = {{DINOv3}},
	doi = {10.48550/arXiv.2508.10104},
	author = {Siméoni, Oriane and Vo, Huy V. and Seitzer, Maximilian and Baldassarre, Federico and Oquab, Maxime and Jose, Cijo and Khalidov, Vasil and Szafraniec, Marc and Yi, Seungeun and Ramamonjisoa, Michaël and Massa, Francisco and Haziza, Daniel and Wehrstedt, Luca and Wang, Jianyuan and Darcet, Timothée and Moutakanni, Théo and Sentana, Leonel and Roberts, Claire and Vedaldi, Andrea and Tolan, Jamie and Brandt, John and Couprie, Camille and Mairal, Julien and Jégou, Hervé and Labatut, Patrick and Bojanowski, Piotr},
	month = aug,
	year = {2025},
	journal = {arXiv preprint arXiv:2508.10104},
}

@inproceedings{wu_rethinking_2021,
	title = {Rethinking and improving relative position encoding for vision transformer},
	booktitle = {Proceedings of the {IEEE}/{CVF} international conference on computer vision},
	author = {Wu, Kan and Peng, Houwen and Chen, Minghao and Fu, Jianlong and Chao, Hongyang},
	year = {2021},
	pages = {10033--10041},
}

@article{chen_jigsaw-vit_2023,
	title = {Jigsaw-{ViT}: {Learning} jigsaw puzzles in vision transformer},
	volume = {166},
	journal = {Pattern Recognition Letters},
	author = {Chen, Yingyi and Shen, Xi and Liu, Yahui and Tao, Qinghua and Suykens, Johan AK},
	year = {2023},
	pages = {53--60},
}

@inproceedings{press_train_2022,
	title = {Train {Short}, {Test} {Long}: {Attention} with {Linear} {Biases} {Enables} {Input} {Length} {Extrapolation}},
	booktitle = {International {Conference} on {Learning} {Representations}},
	author = {Press, Ofir and Smith, Noah A and Lewis, Mike},
	year = {2022},
}

@inproceedings{ren_masked_2023,
	title = {Masked jigsaw puzzle: {A} versatile position embedding for vision transformers},
	booktitle = {Proceedings of the {IEEE}/{CVF} {Conference} on {Computer} {Vision} and {Pattern} {Recognition}},
	author = {Ren, Bin and Liu, Yahui and Song, Yue and Bi, Wei and Cucchiara, Rita and Sebe, Nicu and Wang, Wei},
	year = {2023},
	pages = {20382--20391},

}

@inproceedings{heo_rotary_2024,
	title = {Rotary position embedding for vision transformer},
	booktitle = {European {Conference} on {Computer} {Vision}},
	publisher = {Springer},
	author = {Heo, Byeongho and Park, Song and Han, Dongyoon and Yun, Sangdoo},
	year = {2024},
	pages = {289--305},
}

@inproceedings{chu_conditional_2021,
	title = {Conditional {Positional} {Encodings} for {Vision} {Transformers}},
	booktitle = {The {Eleventh} {International} {Conference} on {Learning} {Representations}},
	author = {Chu, Xiangxiang and Tian, Zhi and Zhang, Bo and Wang, Xinlong and Shen, Chunhua},
	year = {2021},
}

@inproceedings{dosovitskiy_image_2020,
	title = {An {Image} is {Worth} 16x16 {Words}: {Transformers} for {Image} {Recognition} at {Scale}},
	booktitle = {International {Conference} on {Learning} {Representations}},
	author = {Dosovitskiy, Alexey and Beyer, Lucas and Kolesnikov, Alexander and Weissenborn, Dirk and Zhai, Xiaohua and Unterthiner, Thomas and Dehghani, Mostafa and Minderer, Matthias and Heigold, Georg and Gelly, Sylvain and Uszkoreit, Jakob and Houlsby, Neil},
	year = {2020},
}

@inproceedings{xu_e_2023,
	title = {E (2)-{Equivariant} {Vision} {Transformer}},
	booktitle = {39th {Conference} on {Uncertainty} in {Artificial} {Intelligence}},
	publisher = {PMLR},
	author = {Xu, Renjun and Yang, Kaifan and Liu, Ke and He, Fengxiang},
	year = {2023},
	pages = {2356--2366},
}

@inproceedings{zhang_positional_2023,
	title = {Positional label for self-supervised vision transformer},
	volume = {37},
	booktitle = {Proceedings of the {AAAI} {Conference} on {Artificial} {Intelligence}},
	author = {Zhang, Zhemin and Gong, Xun},
	year = {2023},
	note = {Issue: 3},
	pages = {3516--3524},
}

@inproceedings{touvron_training_2021,
	title = {Training data-efficient image transformers \& distillation through attention},
	booktitle = {International conference on machine learning},
	publisher = {PMLR},
	author = {Touvron, Hugo and Cord, Matthieu and Douze, Matthijs and Massa, Francisco and Sablayrolles, Alexandre and Jégou, Hervé},
	year = {2021},
	pages = {10347--10357},
}

@inproceedings{haviv_transformer_2022,
	title = {Transformer {Language} {Models} without {Positional} {Encodings} {Still} {Learn} {Positional} {Information}},
	booktitle = {Findings of the {Association} for {Computational} {Linguistics}: {EMNLP} 2022},
	author = {Haviv, Adi and Ram, Ori and Press, Ofir and Izsak, Peter and Levy, Omer},
	year = {2022},
	pages = {1382--1390},
}

@article{jiang_encoding_2022,
	title = {The encoding method of position embeddings in vision transformer},
	volume = {89},
	doi = {10.1016/j.jvcir.2022.103664},
	journal = {Journal of Visual Communication and Image Representation},
	author = {Jiang, Kai and Peng, Peng and Lian, Youzao and Xu, Weisheng},
	year = {2022},
	pages = {Article 103664},
}

@article{han_survey_2022,
	title = {A survey on vision transformer},
	volume = {45},
	number = {1},
	doi = {10.1109/TPAMI.2022.3152247},
	journal = {IEEE Transactions on Pattern Analysis and Machine Intelligence},
	author = {Han, Kai and Wang, Yunhe and Chen, Hanting and Chen, Xinghao and Guo, Jianyuan and Liu, Zhenhua and Tang, Yehui and Xiao, An and Xu, Chunjing and Xu, Yixing and Yang, Zhaohui and Tao, Dacheng},
	year = {2022},
	pages = {87--110},
}

@inproceedings{liu_swin_2022,
	title = {Swin transformer v2: {Scaling} up capacity and resolution},
	booktitle = {Proceedings of the {IEEE}/{CVF} conference on computer vision and pattern recognition},
	author = {Liu, Ze and Hu, Han and Lin, Yutong and Yao, Zhuliang and Xie, Zhenda and Wei, Yixuan and Ning, Jia and Cao, Yue and Zhang, Zheng and Dong, Li and Wei, Furu and Guo, Baining},
	year = {2022},
	pages = {12009--12019},

}

@inproceedings{xiao_unified_2018,
	title = {Unified perceptual parsing for scene understanding},
	booktitle = {Proceedings of the {European} conference on computer vision ({ECCV})},
	author = {Xiao, Tete and Liu, Yingcheng and Zhou, Bolei and Jiang, Yuning and Sun, Jian},
	year = {2018},
	pages = {418--434},
}

@article{yang2024depth,
  title={Depth anything v2},
  author={Yang, Lihe and Kang, Bingyi and Huang, Zilong and Zhao, Zhen and Xu, Xiaogang and Feng, Jiashi and Zhao, Hengshuang},
  journal={Advances in Neural Information Processing Systems},
  volume={37},
  pages={21875--21911},
  year={2024}
}

@inproceedings{raghu_vision_2021,
	title = {Do {Vision} {Transformers} {See} {Like} {Convolutional} {Neural} {Networks}?},
	volume = {34},
	booktitle = {Advances in {Neural} {Information} {Processing} {Systems}},
	publisher = {Curran Associates, Inc.},
	author = {Raghu, Maithra and Unterthiner, Thomas and Kornblith, Simon and Zhang, Chiyuan and Dosovitskiy, Alexey},
	editor = {Ranzato, M. and Beygelzimer, A. and Dauphin, Y. and Liang, P. S. and Vaughan, J. Wortman},
	year = {2021},
	pages = {12116--12128},
}

@inproceedings{doersch_unsupervised_2015,
	title = {Unsupervised {Visual} {Representation} {Learning} by {Context} {Prediction}},
	booktitle = {International {Conference} on {Computer} {Vision} ({ICCV})},
	author = {Doersch, Carl and Gupta, Abhinav and Efros, Alexei A.},
	year = {2015},
}

@inproceedings{naseer_intriguing_2021,
	address = {Red Hook, NY, USA},
	series = {{NIPS} '21},
	title = {Intriguing properties of vision transformers},
	isbn = {978-1-7138-4539-3},
	booktitle = {Proceedings of the 35th {International} {Conference} on {Neural} {Information} {Processing} {Systems}},
	publisher = {Curran Associates Inc.},
	author = {Naseer, Muzammal and Ranasinghe, Kanchana and Khan, Salman and Hayat, Munawar and Khan, Fahad Shahbaz and Yang, Ming-Hsuan},
	year = {2021},
}

@inproceedings{xiao_early_2021,
	address = {Red Hook, NY, USA},
	series = {{NIPS} '21},
	title = {Early convolutions help transformers see better},
	isbn = {978-1-7138-4539-3},
	booktitle = {Proceedings of the 35th {International} {Conference} on {Neural} {Information} {Processing} {Systems}},
	publisher = {Curran Associates Inc.},
	author = {Xiao, Tete and Singh, Mannat and Mintun, Eric and Darrell, Trevor and Dollár, Piotr and Girshick, Ross},
	year = {2021},
	
}

@article{kim_solving_2025,
	title = {Solving jigsaw puzzles by predicting fragment’s coordinate based on vision transformer},
	volume = {272},
	issn = {0957-4174},
	doi = {10.1016/j.eswa.2025.126776},
	journal = {Expert Systems with Applications},
	author = {Kim, Garam and Cho, Hyeonseong and Nam, Hyoungsik},
	month = may,
	year = {2025},
	pages = {Article 126776},
}

@inproceedings{paszke_pytorch_2019,
	title = {{PyTorch}: An imperative style, high-performance deep learning library},
	author = {Paszke, Adam and Gross, Sam and Massa, Francisco and Lerer, Adam and Bradbury, James and Chanan, Gregory and Killeen, Trevor and Lin, Zeming and Gimelshein, Natalia and Antiga, Luca and Desmaison, Alban and Kopf, Andreas and Yang, Edward and DeVito, Zachary and Raison, Martin and Tejani, Alykhan and Chilamkurthy, Sasank and Steiner, Benoit and Fang, Lu and Bai, Junjie and Chintala, Soumith},
	booktitle = {Advances in Neural Information Processing Systems},
	volume = {32},
	pages = {8024--8035},
	year = {2019},
}

@inproceedings{loshchilov_decoupled_2019,
	title = {Decoupled weight decay regularization},
	author = {Loshchilov, Ilya and Hutter, Frank},
	booktitle = {International Conference on Learning Representations},
	year = {2019},
}

@inproceedings{deng_imagenet_2009,
	title = {{ImageNet}: A large-scale hierarchical image database},
	author = {Deng, Jia and Dong, Wei and Socher, Richard and Li, Li-Jia and Li, Kai and Fei-Fei, Li},
	booktitle = {2009 IEEE Conference on Computer Vision and Pattern Recognition},
	pages = {248--255},
	year = {2009},
	doi = {10.1109/CVPR.2009.5206848},
}

@inproceedings{zhou_scene_2017,
	title = {Scene parsing through {ADE20K} dataset},
	author = {Zhou, Bolei and Zhao, Hang and Puig, Xavier and Fidler, Sanja and Barriuso, Adela and Torralba, Antonio},
	booktitle = {2017 IEEE Conference on Computer Vision and Pattern Recognition},
	pages = {5122--5130},
	year = {2017},
	doi = {10.1109/CVPR.2017.544},
}

@inproceedings{roberts_hypersim_2021,
	title = {{Hypersim}: A photorealistic synthetic dataset for holistic indoor scene understanding},
	author = {Roberts, Mike and Ramapuram, Jason and Ranjan, Anurag and Kumar, Atulit and Bautista, Miguel Angel and Paczan, Nathan and Webb, Russ and Susskind, Joshua M.},
	booktitle = {2021 IEEE/CVF International Conference on Computer Vision},
	pages = {10892--10902},
	year = {2021},
	doi = {10.1109/ICCV48922.2021.01073},
}

@article{tang_bio-inspired_2023,
	title = {A bio-inspired positional embedding network for transformer-based models},
	volume = {166},
	issn = {0893-6080},
	doi = {10.1016/j.neunet.2023.07.015},
	journal = {Neural Networks},
	author = {Tang, Xue-song and Hao, Kuangrong and Wei, Hui},
	year = {2023},
	pages = {204--214},
}

@article{su_dctvit_2024,
	title = {{DctViT}: Discrete Cosine Transform meet vision transformers},
	volume = {172},
	issn = {0893-6080},
	doi = {10.1016/j.neunet.2024.106139},
	journal = {Neural Networks},
	author = {Su, Keke and Cao, Lihua and Zhao, Botong and Li, Ning and Wu, Di and Han, Xiyu and Liu, Yangfan},
	year = {2024},
	pages = {Article 106139},
}

@article{zhu_central_2025,
	title = {Central loss guides coordinated {Transformer} for reliable anatomical landmark detection},
	volume = {187},
	issn = {0893-6080},
	doi = {10.1016/j.neunet.2025.107391},
	journal = {Neural Networks},
	author = {Zhu, Qikui and Bi, Yihui and Chen, Jie and Chu, Xiangpeng and Wang, Danxin and Wang, Yanqing},
	year = {2025},
	pages = {Article 107391},
}

\end{document}